%% file: iclr2026_conference.tex
\documentclass{article} 
\usepackage{iclr2026_conference,times}

\input{math_commands.tex}

\usepackage[utf8]{inputenc}
\usepackage[T1]{fontenc}
\usepackage{url}
\usepackage{booktabs}
\usepackage{amsmath}
\usepackage{amsfonts}
\usepackage{nicefrac}
\usepackage{microtype}
\usepackage{xcolor}
\usepackage{graphicx}
\usepackage{multirow}
\usepackage{pifont}
\usepackage{colortbl}
\usepackage{threeparttable}
\usepackage{xspace}
\usepackage{todonotes}
\usepackage[etex=true,export]{adjustbox}
\usepackage{makecell}
\usepackage{subcaption}
\usepackage{epsfig}
\usepackage{epstopdf}
\usepackage{bbding}
\usepackage{diagbox}
\usepackage{footnote}
\usepackage{wrapfig}
\usepackage[normalem]{ulem}
\usepackage{enumitem}
\usepackage{tikz}
\usepackage{algorithm}
\usepackage{algpseudocode}
\usepackage{textcomp}
\usepackage{chngcntr}

\definecolor{citecolor}{HTML}{0071bc}
\usepackage[colorlinks=true,linkcolor=red,citecolor=citecolor]{hyperref}
\usepackage[capitalise]{cleveref}
\usepackage[table]{xcolor}

\definecolor{revised}{RGB}{0,0,0}

\title{Beyond Passive Imitation: On-Policy Mid-Training Distillation for Language Models}

\makeatletter
\let\original@fnsymbol\@fnsymbol
\renewcommand{\@fnsymbol}[1]{%
  \ifnum#1=1 \text{\Envelope}\else\original@fnsymbol{#1}\fi}
\makeatother

\author{Ruize Zheng\textsuperscript{1} \quad
  Xuyang Liu\textsuperscript{2} \quad
  Siyu Ren\textsuperscript{3} \quad
  Xiaobing Tu\textsuperscript{4} \\
  \textbf{Yinggui Wang}\textsuperscript{4} \quad
  \textbf{Jinkui Ren}\textsuperscript{4} \quad
  \textbf{Xiantao Zhang}\textsuperscript{4} \quad
  \textbf{Linfeng Zhang}\textsuperscript{5}\thanks{Corresponding author.} \\
  \textsuperscript{1}Fudan University \quad
  \textsuperscript{2}The Hong Kong Polytechnic University \\
  \textsuperscript{3}Sichuan University \quad
  \textsuperscript{4}Alibaba Group \quad
  \textsuperscript{5}Shanghai Jiao Tong University \\
  \vspace{5pt}
  \textbf{Code:} \href{https://github.com/ShuidiH20/OPMD}{\textcolor{citecolor}{\textbf{\texttt{https://github.com/ShuidiH20/OPMD}}}}
}

\iclrfinalcopy 
\begin{document}
\maketitle

\input{sections/0_Abstract}
\input{sections/1_Introduction}
\input{sections/2_Related_Work}
\input{sections/3_Methodology}
\input{sections/4_Experiments}
\input{sections/5_Conclusion}
\input{sections/6_Statements}

\bibliography{iclr2026_conference}
\bibliographystyle{iclr2026_conference}

\appendix
\input{sections/X_Appendix}

\end{document}

%% file: math_commands.tex
\usepackage{amsmath,amsfonts,bm}

\def\eqref#1{equation~\ref{#1}}

\def\1{\bm{1}}

\DeclareMathAlphabet{\mathsfit}{\encodingdefault}{\sfdefault}{m}{sl}
\SetMathAlphabet{\mathsfit}{bold}{\encodingdefault}{\sfdefault}{bx}{n}



%% file: sections/0_Abstract.tex
\vspace{-20pt}
\begin{abstract}
\vspace{-5pt}
Continual pre-training (CPT) adapts large language models (LLMs) to specialized domains, yet its standard next-token prediction (NTP) objective treats every corpus token as a fixed target, regardless of its utility to the current model. In a controlled study on the same 1B-token mathematical corpus, we observe sharply different outcomes across base models: NTP improves Llama-3.2-3B's performance on mathematical benchmarks but substantially degrades Qwen3-1.7B-Base's performance, indicating that, under a fixed training recipe, the same training sequences can have model-dependent effects. To address this problem, we introduce \textbf{On-Policy Mid-Training Distillation (OPMD)}, which reformulates mid-training from offline text imitation into active, query-driven knowledge acquisition. At selected anchor positions, OPMD samples trajectories from the evolving student policy and applies dense token-level reverse-KL rewards from a frozen teacher, training the model on states it actually visits rather than only on ground-truth prefixes. This on-policy objective reduces the training--generation mismatch of teacher-forced NTP. On MATH-500 and Minerva, OPMD improves the Qwen student over its base performance and outperforms the evaluated NTP baseline, which degrades on both benchmarks. OPMD integrates into CPT pipelines without added inference-time overhead and offers a model-aware alternative to passive domain adaptation.

\end{abstract}

%% file: sections/1_Introduction.tex
\section{Introduction}
\label{sec:intro}

Continual pre-training (CPT), often referred to as mid-training in recent large language model (LLM) pipelines, has become an effective paradigm for adapting general-purpose LLMs~\citep{grattafiori2024llama,yang2025qwen3,xu2026deepseekv4} to specialized domains such as mathematics and code~\citep{gururangan-etal-2020-dont,ke2023continual}. By continuing autoregressive pre-training on domain-specific corpora, models can acquire substantial domain knowledge while preserving much of their broader capability~\citep{shao2024deepseekmathpushinglimitsmathematical}. In practice, CPT typically continues to optimize the standard next-token prediction (NTP) objective over pre-collected text drawn from the target domain of interest. Under this teacher-forced objective, the model acts as an offline language imitator: it minimizes token-level cross-entropy against the next tokens in fixed, ground-truth sequences during training.

However, standard NTP-based mid-training follows a fundamentally passive learning dynamic. The model is optimized against a fixed, offline corpus without explicitly assessing how informative or useful each training sequence is for its current policy~\citep{ying-etal-2025-data}. Consequently, both the quality and composition of the corpus can substantially affect the resulting model. Recent studies of continual pre-training show that corpus quality, distribution, mixture, and redundancy influence its learning dynamics and downstream performance~\citep{guo-etal-2025-efficient-domain,chen-etal-2025-towards-effective}. These observations expose a central limitation of NTP-based CPT: its effectiveness remains closely tied to which sequences enter the training corpus and how frequently they are sampled.

\begin{figure*}[t]
    \centering
    \includegraphics[width=\linewidth]{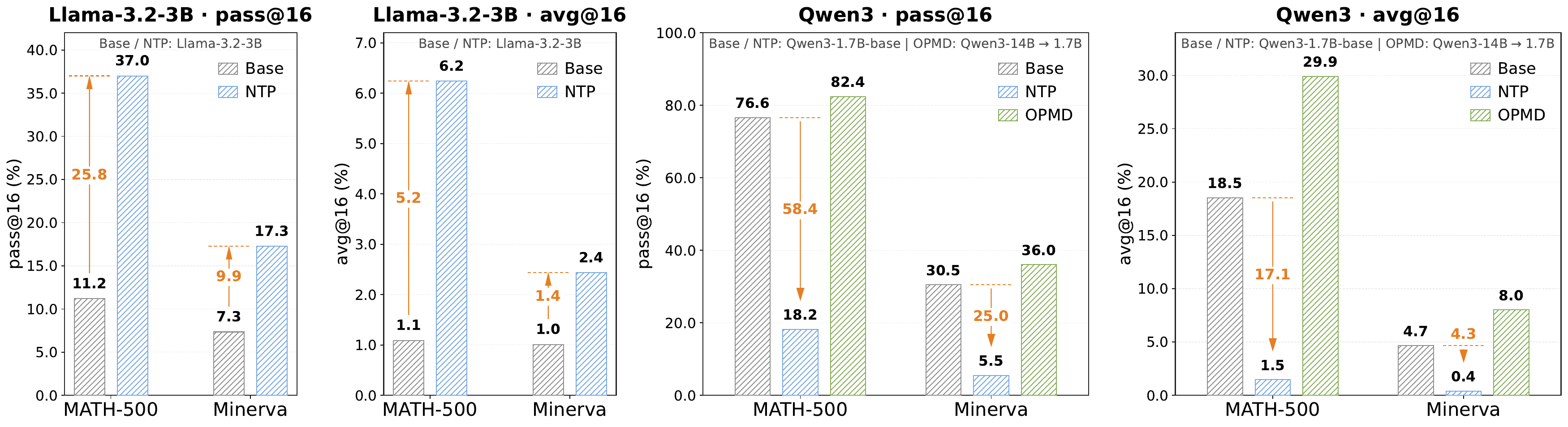}
    \caption{\textbf{Empirical validation of NTP mid-training data sensitivity across base models with different capabilities.} Under identical NTP CPT on a 1B-token subset streamed from MegaMath-Web-Pro, Llama-3.2-3B achieves accuracy gains, whereas Qwen3-1.7B-Base suffers substantial performance degradation on the MATH-500 and Minerva benchmarks.}
    \label{fig:pilot_cpt_degradation}
\end{figure*}

To examine this sensitivity empirically, we conduct a controlled pilot study with two base models that differ in their initial levels of mathematical capability: Llama-3.2-3B~\citep{grattafiori2024llama} and Qwen3-1.7B-Base~\citep{yang2025qwen3}. We apply the same NTP mid-training recipe to both models using a 1B-token subset randomly sampled via streaming from MegaMath-Web-Pro. After that, we evaluate them on MATH-500~\citep{hendrycks2021math500,LightmanMATH500} and Minerva~\citep{lewkowycz2022Minerva}. Figure~\ref{fig:pilot_cpt_degradation} reveals opposite outcomes under the same corpus and training objective: Llama-3.2-3B improves on both benchmarks, whereas Qwen3-1.7B-Base suffers substantial degradation on both. This contrast indicates that the utility of a fixed CPT corpus can depend strongly on the capability and pretraining state of each model.

We hypothesize that a stronger base model has more developed reasoning behaviors to preserve and may therefore be more vulnerable to conflicting or low-utility training signals. Current practice mainly addresses corpus sensitivity through static data curation, including heuristic filtering and mixture reweighting~\citep{NEURIPS2023_dcba6be9,chen-etal-2025-towards-effective}. These methods can improve the training distribution, but they operate before optimization and do not respond to the states that the student model actually visits as training progresses. Teacher-forced NTP introduces a separate limitation: it trains on ground-truth prefixes rather than contexts generated by the model's own policy. This training--generation mismatch produces exposure bias~\citep{pmlr-v15-ross11a,ICLR2024_4db16435}.

Addressing these limitations requires a mid-training objective that can focus learning on informative contexts and supervise the model on its own trajectories. On-Policy Distillation (OPD) offers such a mechanism by sampling trajectory rollouts from the student's evolving policy $\pi_\theta$ and minimizing reverse KL divergence to a teacher policy $\pi_\psi$~\citep{ICLR2024_5be69a58,li2026rethinkingonpolicydistillationlarge,chen2026futurebridge,liu2026specopd}. Because the student receives supervision on states induced by its own decisions during trajectory generation, OPD directly reduces the mismatch between training and generation. Existing OPD methods, however, primarily target post-training settings such as instruction-following alignment and reasoning capability enhancement. Their use during continual pre-training, especially for acquiring domain knowledge without eroding a capable base model, remains underexplored.

To address this gap, we introduce \textbf{On-Policy Mid-Training Distillation (OPMD)}, which reformulates mid-training as an active, query-driven knowledge acquisition process. OPMD first identifies high-uncertainty anchor positions along raw domain sequences and converts each preceding context into a rollout prompt. The student then generates a trajectory under its current policy $\pi_\theta$, while a frozen teacher $\pi_\psi$ provides dense token-level reverse-KL supervision on the states visited by that trajectory. This design moves beyond uniform imitation of corpus tokens by concentrating training on informative contexts and aligning the student where its own policy takes it. OPMD changes only the training procedure, so it can be integrated into existing CPT pipelines without a deployment-time teacher, extra model modules, or additional inference cost at deployment.

We make three contributions to continual pre-training through on-policy distillation:
\begin{enumerate}[leftmargin=10pt, topsep=0pt, itemsep=1pt, parsep=1pt]
    \item \textbf{Controlled empirical finding.} We identify a model-dependent failure mode of passive CPT through a controlled pilot study. With the same 1B-token corpus and NTP training recipe, Llama-3.2-3B improves while Qwen3-1.7B-Base degrades on MATH-500 and Minerva. This contrast establishes that a fixed CPT corpus can have model-dependent utility.
    \item \textbf{On-Policy Mid-Training Distillation.} We formulate \texttt{OPMD} for the continual pre-training stage by combining uncertainty-based anchor selection, rollouts from the evolving student policy, and dense token-level reverse-KL supervision from a frozen teacher. The resulting objective concentrates knowledge transfer on informative contexts and trains the student on states induced by its own generations rather than only on ground-truth corpus prefixes.
    \item \textbf{Evaluation and analysis.} We evaluate \texttt{OPMD} on mathematical benchmarks against an NTP-based CPT baseline and across controlled anchor-selection variants with the same training-sample budget in each arm. The results show that \texttt{OPMD} improves the Qwen student over its base performance while avoiding the degradation observed with NTP mid-training. The anchor-selection comparisons examine how different rollout starting positions relate to these gains.
\end{enumerate}

%% file: sections/2_Related_Work.tex
\section{Related Work}


\noindent\textbf{Continual Pre-Training and Data Curation.} Continual pre-training (CPT), or mid-training, serves as a standard paradigm for infusing domain-specific knowledge—such as mathematics and programming code—into general-purpose base language models~\citep{gururangan-etal-2020-dont, ke2023continual}. In the current literature and industrial practices, the vast majority of CPT approaches strictly adhere to the standard Next Token Prediction (NTP) autoregressive objective over target domain corpora~\citep{lewkowycz2022Minerva, shao2024deepseekmathpushinglimitsmathematical,grattafiori2024llama}. To maximize model performance under the NTP objective, existing research heavily relies on data-centric curation paradigms to enhance corpus quality, which can be broadly categorized into: \textbf{(i)} \textit{Heuristic and deduplication filtering}, which prunes low-quality, noisy, or redundant text using static rules or perplexity thresholds~\citep{NEURIPS2024_370df50c}; and \textbf{(ii)} \textit{Distribution matching and domain reweighting}, which selects target-relevant domain subsets or optimizes domain mixture ratios prior to training~\citep{NEURIPS2023_6b9aa8f4, NEURIPS2023_dcba6be9}. Although these data curation methods effectively enhance corpus purity, they treat data processing as a static, passive pre-processing step independent of the model's active learning dynamics and the contexts it visits. Consequently, they fail to alter the model's rigid reliance on data quality at the level of the training mechanism during mid-training.


\noindent\textbf{On-Policy Distillation in LLMs.} Knowledge Distillation (KD) transfers parametric capabilities and reasoning patterns from massive teacher models to smaller student models~\citep{kim2016sequence, gou2021knowledge,ICLR2024_8ac015d4}. However, traditional KD approaches in LLMs predominantly operate in an off-policy manner on static datasets or teacher-generated traces~\citep{sanh2019distilbert, hwang2023seqkd}. As a result, the student is trained only in contexts frequented by the teacher rather than in states encountered under its own policy, introducing severe compounding errors during sequence generation: an early mistake made by the student places it in unfamiliar states unobserved during training, causing its policy to diverge rapidly from target behavior~\citep{pmlr-v15-ross11a, ICLR2024_4db16435}. To overcome this discrepancy, On-Policy Distillation (OPD)—exemplified by MiniLLM and Generalized Knowledge Distillation (GKD)—enables student models to sample trajectory rollouts under their own evolving policy $\pi_\theta$ while optimizing reverse KL divergence against a frozen teacher $\pi_\psi$, effectively teaching the student to recover from its own generation mistakes~\citep{ICLR2024_8ac015d4, ICLR2024_5be69a58}. Recent empirical investigations further analyze OPD's underlying mechanics, demonstrating how dense token-level reward assignment effectively stabilizes state exploration and mitigates error accumulation~\citep{li2026rethinkingonpolicydistillationlarge}. Overall, existing OPD methodologies mainly focus on either instruction-following alignment or model compression in task-specific post-training phases. 

Unlike prior CPT strategies that passively adapt corpora to the rigid NTP loss, we propose On-Policy Mid-Training Distillation (OPMD). By introducing on-policy trajectory sampling and optimizing student exploration via token-level reverse KL rewards during continual pre-training, OPMD transforms mid-training into an active, query-driven knowledge acquisition process.

%% file: sections/3_Methodology.tex
\section{Methodology}

\subsection{Preliminaries}
\label{subsec:preliminaries}

In this section, we formally define the problem setup for Continual Pre-Training (CPT) and review the mathematical formulations of On-Policy Distillation (OPD).

\noindent \textbf{Continual Pre-Training via Next Token Prediction.}
Let $\mathcal{D}$ denote a target-domain pre-training corpus consisting of token sequences $\mathbf{x} = (x_1, x_2, \dots, x_T)$, where each token $x_t \in \mathcal{V}$ belongs to a vocabulary $\mathcal{V}$. Standard continual pre-training (CPT) optimizes a base language model parameterized by $\theta$, denoted as policy $\pi_\theta(x_t \mid \mathbf{x}_{<t})$, using the causal Next Token Prediction (NTP) objective. The standard autoregressive cross-entropy loss over corpus $\mathcal{D}$ is formulated as:
\begin{equation}
\mathcal{L}_{\text{NTP}}(\theta) = -\mathbb{E}_{\mathbf{x} \sim \mathcal{D}} \left[ \sum_{t=1}^T \log \pi_\theta(x_t \mid \mathbf{x}_{<t}) \right].
\label{eq:ntp_loss}
\end{equation}
While $\mathcal{L}_{\text{NTP}}$ forces the model to strictly imitate the empirical text distribution of $\mathcal{D}$, it treats all tokens with uniform loss weights regardless of token difficulty or data noise, making optimization highly sensitive to corpus quality.

\noindent \textbf{On-Policy Distillation in LLMs.}
To overcome distribution shift inherent to off-policy sequence imitation, On-Policy Distillation (OPD) leverages a frozen teacher model $\pi_\psi$ to guide a student model $\pi_\theta$. Instead of calculating distillation targets over fixed ground-truth corpora, OPD allows the student model to generate trajectory rollouts sampled directly from its own evolving policy. Formally, using the reverse Kullback-Leibler (KL) divergence to mitigate state-drift and prevent over-generalization, the standard OPD objective is written as:
\begin{equation}
\mathcal{L}_{\text{OPD}}(\theta) = \mathbb{E}_{\mathbf{x} \sim \mathcal{D}, \, \mathbf{y} \sim \pi_\theta(\cdot \mid \mathbf{x})} \left[ \sum_{k=1}^K \mathcal{D}_{\text{KL}}\left( \pi_\theta(\cdot \mid \mathbf{x}, \mathbf{y}_{<k}) \,\parallel\, \pi_\psi(\cdot \mid \mathbf{x}, \mathbf{y}_{<k}) \right) \right],
\label{eq:opd_loss}
\end{equation}
where \(\mathbf{y}=(y_1,\ldots,y_K)\) denotes a trajectory of \(K\) tokens sampled from the student policy, and \(\mathcal{D}_{\mathrm{KL}}(P\parallel Q)=\sum_{v\in\mathcal V}P(v)\log\frac{P(v)}{Q(v)}\) denotes the reverse KL divergence.

\subsection{On-Policy Mid-Training Distillation (OPMD)}
\label{subsec:opmd}

Applying On-Policy Distillation (OPD) to Continual Pre-Training (CPT) offers an appealing framework to overcome the rigid empirical matching of standard Next-Token Prediction (NTP), allowing the model to actively explore trajectory spaces and mitigate distribution shifts. However, directly extending token-level OPD to vast pre-training corpora introduces severe computational bottlenecks: performing autoregressive rollouts and continuous teacher forward passes across billions of unstructured tokens incurs prohibitive compute costs and severe window overlap redundancies. This raises a critical question: \textit{How can we efficiently leverage vast, unstructured CPT corpora for on-policy distillation under bounded computational budgets?}


To address this challenge, we introduce \textit{On-Policy Mid-Training Distillation} (OPMD). OPMD selectively triggers student trajectory rollouts at semantically critical trigger positions via \textit{Anchor Token Identification} (ATI), converts preceding sequence context into prompts, and performs dense, token-level reverse KL distillation against a frozen teacher model.

\begin{figure*}[t]
\centering
\includegraphics[width=\linewidth]{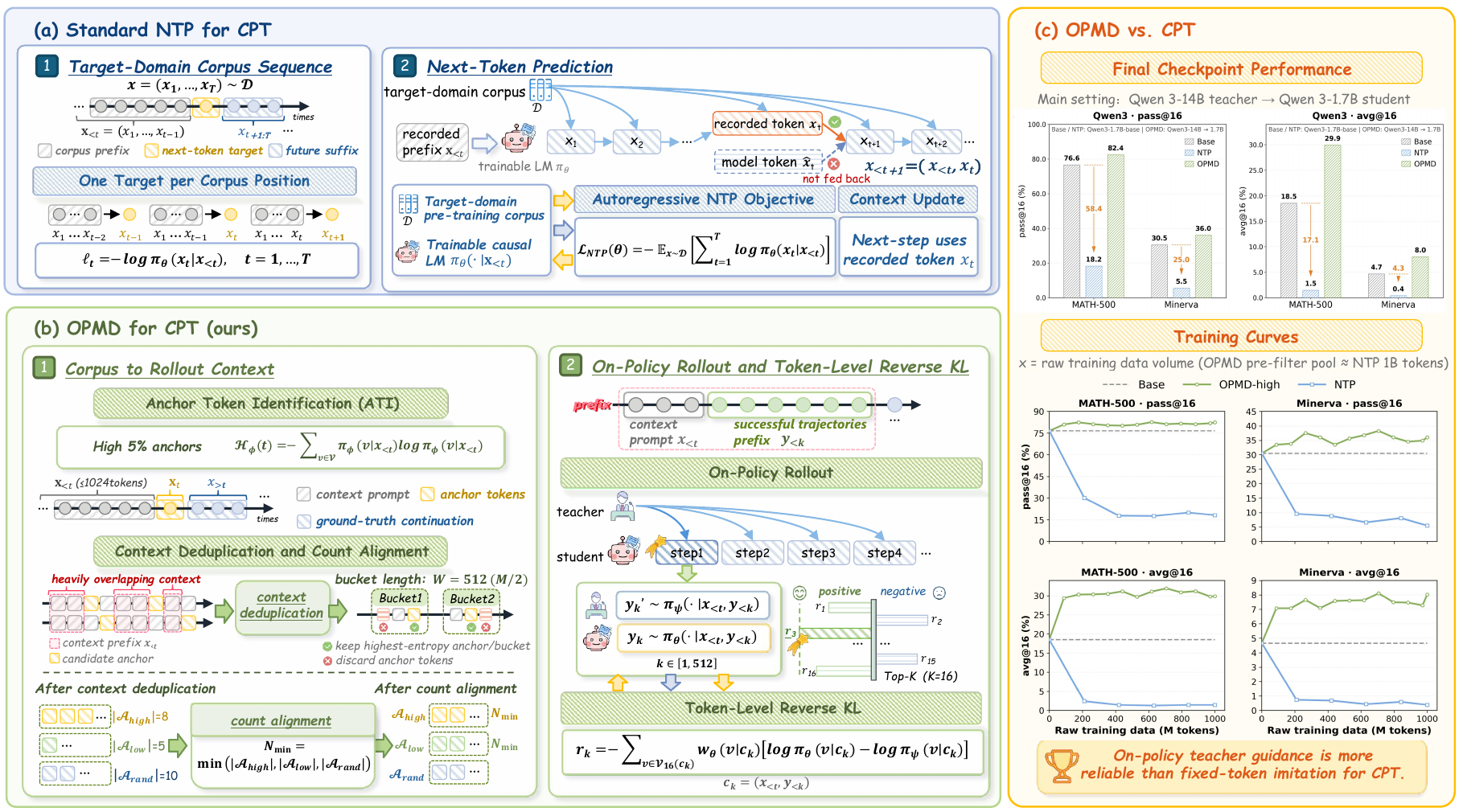}
\caption{\textbf{Overview of the OPMD compared with standard NTP, along with training dynamics.} 
\textbf{(a) Standard NTP:} Rigid token-by-token imitation on specialized domain corpora leads to severe policy collapse. 
\textbf{(b) OPMD Pipeline:} (1) \textit{Anchor Token Identification (ATI) \& Alignment:} A proxy model identifies informative trigger tokens via predictive entropy, followed by context deduplication and strict budget alignment. (2) \textit{On-Policy Rollout \& Distillation:} The student generates continuations from prefix prompts and receives dense teacher guidance through a student-weighted Top-16 log-ratio reward at each rollout step.
\textbf{(c) Performance Dynamics:} OPMD completely eliminates NTP policy collapse and achieves substantial, stable reasoning gains across benchmarks.}
\label{fig:overview}
\end{figure*}

\noindent \textbf{Anchor Token Identification (ATI).}
Not all tokens in a continual pre-training sequence contribute equally to model distillation. Inspired by recent exploration strategies in reinforcement pre-training~\citep{dong2025reinforcementpretraining}, ATI dynamically identifies critical trigger positions for student rollouts based on individual token-level uncertainty. Given an input text sequence $\mathbf{x} = (x_1, \dots, x_T) \sim \mathcal{D}$, we utilize a lightweight, frozen proxy model $\pi_\phi$ (\texttt{DeepSeek-R1-Distill-Qwen-1.5B}) to scan the document. For long sequences exceeding the proxy's context limit, we process text using a sliding window of length 4096 with a stride of 2048, averaging predictive entropies across overlapping regions. The per-token uncertainty at index $t$ is evaluated as:
\begin{equation}
\mathcal{H}_\phi(t) = -\sum_{v \in \mathcal{V}} \pi_\phi(v \mid \mathbf{x}_{<t}) \log \pi_\phi(v \mid \mathbf{x}_{<t}).
\end{equation}

To systematically investigate the effect of token uncertainty on mid-training distillation, we construct three distinct experimental arms based on per-token entropy distributions:
\begin{itemize}
    \item \textit{High-Entropy Arm:} We target positions with maximum predictive uncertainty to encourage active policy exploration, extracting the top 5\% ($r = 0.05$) highest-entropy tokens per document:
    \begin{equation}
    \mathcal{S}_{\text{high}}(t) = \mathbb{I}\left( \mathcal{H}_\phi(t) \ge Q_{0.95}(\mathcal{H}_\phi) \right),
    \end{equation}
    where $Q_{\alpha}(\mathcal{H}_\phi)$ denotes the $\alpha$-quantile of the entropy values within a sequence, and $\mathbb{I}(\cdot)$ is the indicator function.
    
    \item \textit{Low-Entropy Arm:} We extract the bottom 5\% of tokens with the lowest individual entropy values via stable sorting, focusing on deterministic structural transitions:
    \begin{equation}
    \mathcal{S}_{\text{low}}(t) = \mathbb{I}\left( \mathcal{H}_\phi(t) \le Q_{0.05}(\mathcal{H}_\phi) \right).
    \end{equation}

    \item \textit{Stochastic Baseline:} Each token position is independently sampled as a candidate anchor with probability $p=0.05$, using deterministic document-level seeds. The candidate-selection indicator is
    \begin{equation}
    \mathcal{S}_{\text{rand}}(t) \sim \operatorname{Bernoulli}(p).
    \end{equation}
    As described below, these candidates undergo the same deduplication procedure as the other arms: the highest-entropy candidate is retained in each 512-token bucket before count alignment.
\end{itemize}

For each candidate anchor position $t$, the preceding prefix $\mathbf{x}_{<t} = (x_1, \dots, x_{t-1})$ (truncated to a maximum context length $M = 1024$) serves as the conditioning context prompt from which the student generates its subsequent on-policy rollout.

\noindent \textbf{Context Deduplication and Count Alignment.} 
In dense text regions, candidate anchor tokens in close proximity share heavily overlapping context prefixes $\mathbf{x}_{<t}$. Triggering separate student rollouts on overlapping prefixes introduces severe computational redundancy without providing novel exploration signals. To resolve this, we apply a non-overlapping window deduplication procedure: for each candidate set independently, candidate anchors within the same document are partitioned into non-overlapping buckets of length \(W=M/2\). Within each bucket, only the candidate anchor with the highest entropy is retained, while the remaining overlapping anchors are discarded. This entropy-based retention rule is applied independently to all three candidate sets, including the Low-Entropy and Stochastic arms. Thus, the Stochastic arm uses random candidate preselection, but its retained anchors are not uniformly random.

Furthermore, because spatial entropy distributions vary across text regions, deduplication yields different candidate set sizes $|\mathcal{A}_{\text{high}}|$, $|\mathcal{A}_{\text{low}}|$, and $|\mathcal{A}_{\text{rand}}|$. To ensure strict variable control and eliminate sample-size confounding across ablation arms, we uniformly downsample the deduplicated anchor sets without replacement to match the minimum budget across all arms:
\begin{equation}
N_{\min} = \min\left(|\mathcal{A}_{\text{high}}|, |\mathcal{A}_{\text{low}}|, |\mathcal{A}_{\text{rand}}|\right).
\end{equation}
This alignment guarantees that all three arms contain the same number of retained anchor prompts, $N_{\min}$.

\noindent \textbf{On-Policy Rollout and Token-Level Reverse KL.}
For each validated anchor position \(t\in\mathcal{A}(\mathbf{x})\), the preceding text \(\mathbf{x}_{<t}\), truncated to the maximum context length \(M=1024\), serves as the prompt for an on-policy rollout. The student generates a single continuation \(\mathbf{y}=(y_1,\ldots,y_K)\) of up to \(K=512\) tokens:
\begin{equation}
y_k\sim\pi_\theta\left(\cdot\mid\mathbf{x}_{<t},\mathbf{y}_{<k}\right),
\qquad k=1,\ldots,K.
\end{equation}
The frozen teacher \(\pi_\psi\) then evaluates the prefixes visited by the student. Thus, distillation guidance is provided on student-generated continuations rather than exclusively on the fixed token sequences in the pre-training corpus.

At each rollout step \(k\), let \(c_k=(\mathbf{x}_{<t},\mathbf{y}_{<k})\), and let \(\mathcal{V}_{16}(c_k)\) denote the student's Top-16 next-token candidates. We assign each candidate a weight proportional to its student probability,
\begin{equation}
w_\theta(v\mid c_k)
=
\frac{\pi_\theta(v\mid c_k)}
{\sum_{u\in\mathcal{V}_{16}(c_k)}\pi_\theta(u\mid c_k)}.
\end{equation}
The token-level distillation reward aggregates the teacher–student log-probability differences over these candidates:
\begin{equation}
r_k
=
-\sum_{v\in\mathcal{V}_{16}(c_k)}
w_\theta(v\mid c_k)
\left[
\log\pi_\theta(v\mid c_k)
-
\log\pi_\psi(v\mid c_k)
\right].
\label{eq:opmd_reward}
\end{equation}
This reward retains the student-weighted log ratio underlying reverse KL while restricting the comparison to a small set of high-probability candidates. The student probabilities are normalized within the Top-16 set, while the teacher log-probabilities retain their full-vocabulary values. It therefore provides a Top-16 approximation to the full-vocabulary distillation signal.

\noindent \textbf{Overall Training Objective.}
Given the anchor prompts and student-generated continuations, OPMD aims to maximize the accumulated teacher-guided rewards. We express this intended distillation objective as
\begin{equation}
\mathcal{L}_{\mathrm{OPMD}}(\theta)
=
-\mathbb{E}_{\substack{
\mathbf{x}\sim\mathcal{D},\;
t\sim\mathcal{A}(\mathbf{x}),\\
\mathbf{y}\sim\pi_\theta(\cdot\mid\mathbf{x}_{<t})
}}
\left[
\sum_{k=1}^{K}r_k
\right].
\label{eq:opmd_loss}
\end{equation}
Here, the expectation covers corpus sequences, selected anchor positions, and continuations sampled from the evolving student policy. In implementation, we use the individual candidate contributions in Eq. (10), rather than the aggregated scalar \(r_k\), as direct advantage signals within the veRL PPO framework. Equation~(\ref{eq:opmd_loss}) states the intended distillation target; PPO supplies the practical update mechanism. This allows OPMD to train on student-visited contexts using dense teacher feedback without outcome rewards or external verifiers.

%% file: sections/4_Experiments.tex
\section{Experiments}
\label{sec:experiments}

\subsection{Experimental Settings}

\noindent \textbf{Training Data.} We perform mid-training on a 1B-token subset of MegaMath-Web-Pro~\citep{zhou2025megamathpushinglimitsopen}. Per-token predictive uncertainty is calculated via a frozen proxy model (\texttt{DeepSeek-R1-Distill-Qwen-1.5B}~\citep{guo2025deepseek}) to construct three ablation arms: High-Entropy (top 5\%), Low-Entropy (bottom 5\%), and Stochastic Baseline (5\% random candidate preselection followed by entropy-based deduplication). Each anchor token defines a preceding context prompt of up to 1024 tokens, targeting an on-policy rollout of up to 512 tokens. Following 512-token window deduplication, all arms are uniformly downsampled to an identical budget of 435,049 training sequences to ensure strict variable control across the three anchor-selection strategies.

\noindent \textbf{Implementation Details.} Built on \texttt{veRL}~\citep{10.1145/3689031.3696075} with \texttt{vLLM}~\citep{10.1145/3600006.3613165}, we employ \texttt{Qwen3-1.7B-Base}~\citep{yang2025qwen3} as the student model and a frozen \texttt{Qwen3-14B-Base}~\citep{yang2025qwen3} as the teacher. The student generates a single trajectory per prompt ($n=1$, temperature 1.0), supervised by a student-weighted Top-16 log-ratio reward and a reference KL penalty ($\beta=0.005$). All OPMD runs use the same training hyperparameters: one epoch of AdamW optimization with a learning rate of $1\times10^{-6}$ on 4--8 NVIDIA A100/A800 GPUs. The NTP baselines use a learning rate of $1\times10^{-5}$. Additional training and implementation details are provided in Appendix~\ref{app:implementation}.

\noindent \textbf{Benchmarks.} We evaluate models on mathematical reasoning tasks across MATH-500~\citep{hendrycks2021math500,LightmanMATH500} and Minerva~\citep{lewkowycz2022Minerva}. For each problem, we sample 16 candidate responses using stochastic decoding and report \textbf{pass@16} and \textbf{avg@16} via rule-based boxed answer matching. The first metric measures whether any of the 16 samples is correct, whereas the second reflects the average correctness of the sampled responses.

\subsection{Main Results}
\label{subsec:main_results}

\input{tables/main_results}

We evaluate the effectiveness of On-Policy Mid-Training Distillation (OPMD) on mathematical reasoning benchmarks across various training paradigms, teacher model capacities, and intermediate checkpoints. Our empirical results highlight three key advantages:

\noindent \textbf{(i) Superiority Over Standard NTP Training.} As summarized in Table~\ref{tab:main_results1}, standard Next-Token Prediction (NTP) suffers from substantial performance degradation on both math benchmarks during continual pre-training on domain-specific mathematical text. Throughout CPT, the NTP baseline's performance drops precipitously from the base model's initial 76.6\% pass@16 down to 18.2\% on MATH-500 (with avg@16 collapsing from 18.5\% to 1.5\%) and from 30.5\% down to 5.5\% pass@16 on Minerva (with avg@16 dropping from 4.7\% to 0.4\%) in our evaluation.

In stark contrast, OPMD improves benchmark performance under the evaluated training settings and unlocks substantial reasoning gains across both teacher configurations. Specifically, distilling with the compact \texttt{Qwen3-4B} teacher (Teacher 1) achieves the highest solve rates on MATH-500, pushing pass@16 to 82.8\% (+6.2\%) and avg@16 to 30.5\% (+12.0\%). Meanwhile, scaling to the \texttt{Qwen3-14B} teacher (Teacher 2) yields the strongest capacity on Minerva, reaching 36.0\% pass@16 (+5.5\%) and 8.0\% avg@16 (+3.4\%). The two metrics reveal different aspects of this improvement.

For example, with the 4B teacher on MATH-500, pass@16 increases by 6.2 percentage points relative to the base student, whereas avg@16 increases by 12.0 points. Thus, the gain is reflected not only in the set of problems solved at least once among 16 samples, but also in the frequency of correct responses across those samples. Overall, all OPMD variants consistently outperform the baseline student and avoid the degradation observed for our NTP baseline.

\begin{figure*}[t]
\centering
\includegraphics[width=\linewidth]{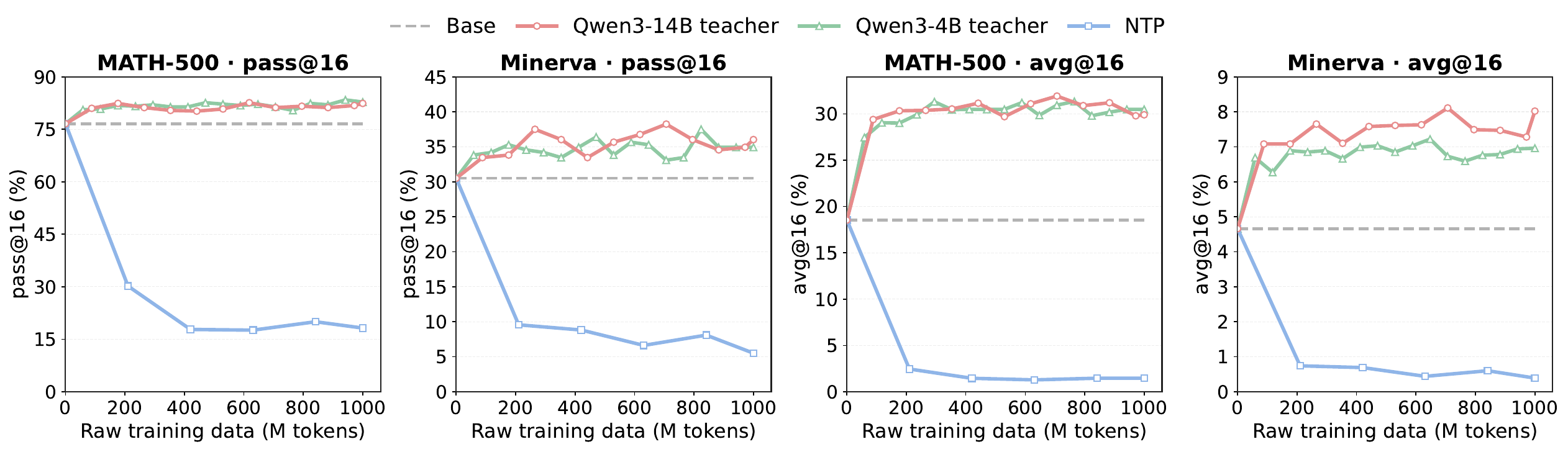}
\caption{\textbf{Training dynamics of OPMD with different teacher scales versus standard NTP.} Performance trajectories (\texttt{pass@16} and \texttt{avg@16}) of the \texttt{Qwen3-1.7B} student on MATH-500 and Minerva are plotted against raw training token volume (in millions). NTP performance declines, whereas OPMD improves with either the Qwen3-4B or Qwen3-14B teacher. The two teacher settings show broadly comparable trajectories over the evaluated range.}
\label{fig:teacher_scale_dynamics}
\end{figure*}

\noindent \textbf{(ii) Scalability and Teacher Capacity Dynamics.}

OPMD demonstrates notable scalability across teacher model capacities ($\text{Qwen3-4B-Base}$ vs. $\text{Qwen3-14B-Base}$). As depicted in Figure~\ref{fig:teacher_scale_dynamics}, distilling the \texttt{Qwen3-1.7B} student with a compact 4B teacher yields closely comparable performance trajectories to those guided by the substantially larger 14B teacher across the entire training token regime (e.g., final MATH-500 pass@16 of 82.8\% for 4B vs. 82.4\% for 14B; final Minerva pass@16 of 34.9\% for 4B vs. 36.0\% for 14B). Note that while total step counts differ due to hardware cluster configurations (1699 steps on 4 GPUs for 4B vs. 1132 steps on 6 GPUs for 14B), both settings reflect an equivalent overall token exposure. Two possible explanations for this plateau under matched token exposure are:
\begin{itemize}
    \item \textit{Student Capacity Ceiling:} The limited parameter capacity of a 1.7B student may constrain its ability to assimilate increasingly granular soft-target nuances beyond a certain threshold. The log-probabilities supplied by a 4B teacher for the top-16 candidates appear sufficient to provide effective directional supervision for a student of this scale, as reflected in the similar teacher-guided trajectories on both evaluated benchmarks. This offers a capacity-based explanation for the small performance differences between the two teacher settings.
    \item \textit{State-Space Distributional Alignment:} Because trajectory rollouts are generated on-policy by the student ($\mathbf{y} \sim \pi_\theta$), evaluated states remain anchored within the student's policy distribution. Over these visited contexts, the 4B and 14B teachers may assign similar relative log-probabilities to influential candidates in the student's Top-16 set, yielding broadly congruent Reverse KL update directions. Similar supervision distributions imply similar updates on the contexts that the student actually visits.
    
\end{itemize}
These empirical findings underscore a practical benefit of OPMD regarding \textbf{teacher compute efficiency}: strong mid-training distillation gains can be realized with a relatively compact teacher, substantially reducing inference and memory overhead during optimization without meaningfully sacrificing target performance for the Qwen3-1.7B student on the two tested benchmarks.

\noindent \textbf{(iii) Exceptional Sample Efficiency and Training Stability.} Across all OPMD variants, the student model achieves remarkable data efficiency, capturing substantial distillation gains within bounded optimization steps. With the 14B teacher, the student's MATH-500 avg@16 rises from 18.5\% at initialization to 29.4\% at step 100, the first evaluated checkpoint, while Minerva avg@16 rises from 4.7\% to 7.1\%. These gains persist despite fluctuations at later checkpoints: at the final checkpoint of this training run, avg@16 reaches 29.9\% on MATH-500 and 8.0\% on Minerva (Appendix Table~\ref{tab:opmd_step_results}). The corresponding pass@16 scores also remain above the base model throughout the evaluated run. These results suggest that on-policy distillation over anchor-triggered contexts can sustain its early improvements throughout the evaluated training budget, including at later checkpoints.

\subsection{Ablation Studies and Analysis}
\label{subsec:ablation}

\noindent \textbf{Effectiveness of Different ATI Strategies.}

\begin{figure*}[t]
    \centering
    \includegraphics[width=\linewidth]{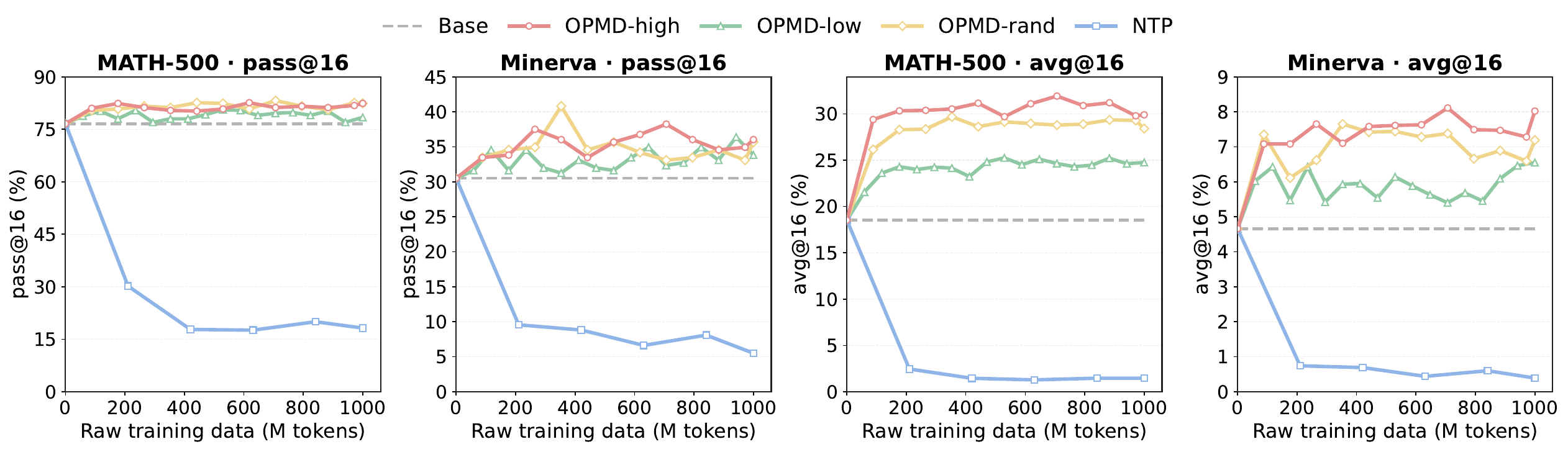}
\caption{\textbf{Training dynamics of Anchor Token Identification (ATI) strategies.}
MATH-500 and Minerva \texttt{pass@16} and \texttt{avg@16} for the Qwen3-1.7B student across all three ATI strategies are plotted against raw training token volume (millions). In the evaluated setting, OPMD variants remain above the base model throughout training, whereas the NTP baseline degrades rapidly. Among these arms, OPMD-high attains the highest observed peak \texttt{avg@16} on both benchmarks.}
\label{fig:ati_trajectories}
\end{figure*}

To isolate the effect of anchor token selection quality, all experimental arms are strictly controlled under an identical training budget of 435,049 samples, aligned via context-aware deduplication over non-overlapping 512-token windows. We systematically evaluate three selection strategies: High-Entropy, Low-Entropy, and Stochastic Baseline (Rand). As illustrated in Figure~\ref{fig:ati_trajectories}, empirical results yield three key insights about peak solution density, low-entropy anchors, and performance under random preselection. Because all arms use the same sample budget, the comparisons isolate the effect of where rollouts begin rather than differences in training volume:
\begin{itemize}
    \item \textit{High-Entropy Candidate Preselection and Peak Solution Density:} Distilling over high-entropy token positions achieves the highest peak solution consistency (\textit{avg@16}) across benchmarks, reaching \textbf{31.9\%} avg@16 on MATH-500 and \textbf{8.1\%} avg@16 on Minerva (both at step 800). High entropy appears to mark local epistemic uncertainty and critical logical branching points in pre-training corpora. Triggering on-policy student rollouts at these decision junctions provides rich teacher supervision, contributing to the higher avg@16 observed in this setting. The peak values show that high-entropy anchors improve solution consistency on both benchmarks while using the same sample budget as the other strategies.
    \item \textit{Limitations of Low-Entropy Anchors:} Conversely, the Low-Entropy arm yields lower overall gains among OPMD variants, peaking at 25.3\% avg@16 on MATH-500 (step 900) and 6.6\% on Minerva (step 1699). Low-entropy positions predominantly correspond to highly deterministic structural or syntactic continuations where the next token is easily predictable. Distilling over such confident context offers limited active exploration value to the student policy. The lower peaks on both benchmarks are consistent with this limitation of low-entropy anchors, with the same ordering observed on MATH-500 and Minerva.
    \item \textit{Framework Robustness and Stochastic Baseline Performance:} Remarkably, the Stochastic (Rand) baseline demonstrates strong instance-level exploration, achieving peak pass@16 scores of \textbf{83.2\%} on MATH-500 (step 800) and \textbf{40.8\%} on Minerva (step 400). This indicates that the core OPMD paradigm---converting static CPT context into active prompt-rollout triggers---is inherently robust to anchor selection. Nevertheless, High-Entropy selection generally exhibits superior average solve rate density (\textit{avg@16}), suggesting greater trajectory stability over extended training. The two metrics thus distinguish the stochastic arm's peak pass@16 from the high-entropy arm's stronger avg@16.
\end{itemize}

%% file: tables/main_results.tex
\definecolor{bestblue}{RGB}{218,232,252}
\definecolor{secondred}{RGB}{255,226,226}
\definecolor{sectiongray}{RGB}{240,240,240}
\definecolor{upgreen}{RGB}{46,139,87}

\definecolor{ntpblue}{HTML}{EEF5FB}
\definecolor{opmdgreen}{HTML}{F2F8EC}
\definecolor{opmddarkgreen}{RGB}{0,128,0}
\definecolor{opmdred}{RGB}{211,47,47}

\begin{table*}[t]
\centering
\caption{\textbf{Main results on MATH-500 and Minerva}
($n{=}16$ sampling). Highlighted values and arrows indicate
performance gains or losses relative to the
\texttt{Qwen3-1.7B-Base} student model.}
\label{tab:main_results1}

\begingroup
\setlength{\tabcolsep}{12pt}
\renewcommand{\arraystretch}{1.05}

\begin{tabular}{lcccc}
\toprule
\multirow{2}{*}{\textbf{Methods}}
& \multicolumn{2}{c}{\textbf{MATH-500}}
& \multicolumn{2}{c}{\textbf{Minerva}} \\
\cmidrule(lr){2-3}
\cmidrule(lr){4-5}
& \textbf{pass@16}
& \textbf{avg@16}
& \textbf{pass@16}
& \textbf{avg@16} \\
\midrule

Qwen3-1.7B-Base (Student)
& 76.6
& 18.5
& 30.5
& 4.7 \\

Qwen3-4B-Base (Teacher 1)
& 84.8
& 21.3
& 34.9
& 5.6 \\

Qwen3-14B-Base (Teacher 2)
& 94.2
& 68.6
& 51.5
& 24.0 \\

\rowcolor{ntpblue}
\textbf{Student NTP}
& $\mathbf{18.2}^{\textcolor{opmddarkgreen}{\downarrow 58.4}}$
& $\mathbf{1.5}^{\textcolor{opmddarkgreen}{\downarrow 17.0}}$
& $\mathbf{5.5}^{\textcolor{opmddarkgreen}{\downarrow 25.0}}$
& $\mathbf{0.4}^{\textcolor{opmddarkgreen}{\downarrow 4.3}}$ \\

\rowcolor{opmdgreen}
\textbf{Teacher 1 OPMD}
& $\mathbf{82.8}^{\textcolor{opmdred}{\uparrow 6.2}}$
& $\mathbf{30.5}^{\textcolor{opmdred}{\uparrow 12.0}}$
& $\mathbf{34.9}^{\textcolor{opmdred}{\uparrow 4.4}}$
& $\mathbf{7.0}^{\textcolor{opmdred}{\uparrow 2.3}}$ \\

\rowcolor{opmdgreen}
\textbf{Teacher 2 OPMD}
& $\mathbf{82.4}^{\textcolor{opmdred}{\uparrow 5.8}}$
& $\mathbf{29.9}^{\textcolor{opmdred}{\uparrow 11.4}}$
& $\mathbf{36.0}^{\textcolor{opmdred}{\uparrow 5.5}}$
& $\mathbf{8.0}^{\textcolor{opmdred}{\uparrow 3.4}}$ \\

\bottomrule
\end{tabular}

\endgroup
\end{table*}

%% file: sections/5_Conclusion.tex
\section{Conclusion}

We studied continual pre-training on a fixed mathematical corpus and observed that the same NTP training recipe improves Llama-3.2-3B but degrades Qwen3-1.7B-Base on the evaluated reasoning benchmarks. We introduced On-Policy Mid-Training Distillation (OPMD), which turns selected corpus prefixes into rollout prompts and trains on student-generated continuations using dense, token-level feedback from a frozen teacher model. On MATH-500 and Minerva, OPMD avoids the regression observed with NTP and improves the Qwen student over its base model on both pass@16 and avg@16 across the two evaluated teacher configurations. In this setting, 4B and 14B teachers yield comparable gains on both mathematical reasoning benchmarks, while high-entropy candidate preselection achieves the highest observed peak avg@16 among the tested selection procedures. These results suggest that on-policy distillation is a promising objective for mathematical mid-training. Evaluation across additional domains and with uniformly random final anchors will help establish the generality of these findings and isolate the effect of anchor selection.

%% file: sections/6_Statements.tex
\subsection*{AI Use Statement}
We used LLM-based tools to assist with drafting and polishing the manuscript and preparing figures based on our experimental data. The tools did not generate or alter the reported experimental data or numerical results. The authors reviewed and verified the AI-assisted material and take full responsibility for the accuracy and integrity of the paper.

\subsection*{Reproducibility Statement}
Sections~\ref{sec:experiments} and Appendix~\ref{app:implementation} describe the data, anchor construction,
training, and evaluation procedures. Appendix~\ref{app:distillation_details} identifies the public generation and grading scripts, and Appendix~B reports checkpoint results.

%% file: sections/X_Appendix.tex
\clearpage
\begingroup
\renewcommand{\topfraction}{0.95}
\renewcommand{\bottomfraction}{0.90}
\renewcommand{\textfraction}{0.05}
\renewcommand{\floatpagefraction}{0.75}
\section*{Appendix}
\pdfbookmark[0]{Appendix}{memfolder-appendix}
\makeatletter
\let\ocSavedAddtocontents\addtocontents
\renewcommand{\addtocontents}[2]{%
  \def\ocCurrentExtension{#1}\def\ocMainContents{toc}%
  \ifx\ocCurrentExtension\ocMainContents
    \ocSavedAddtocontents{apx}{#2}%
  \else
    \ocSavedAddtocontents{#1}{#2}%
  \fi
}
\setcounter{tocdepth}{2}
\begingroup
\hypersetup{linkcolor=black}
\color{black}
\@starttoc{apx}
\endgroup
\makeatother
\medskip

\section{Additional Implementation Details}
\label{app:implementation}

\subsection{Training Corpus and Anchor Construction}
\label{app:training_data}

All NTP and OPMD experiments use the same sampled set of original
documents from the 1B-token MegaMath-Web-Pro subset described in
Section~4.1. The methods use these documents differently: NTP learns
from the recorded continuations, whereas OPMD converts selected
document prefixes into prompts and trains on student-generated
continuations. Thus, sharing source documents does not imply that
the methods receive identical optimization examples.

For anchor construction, a frozen
\texttt{DeepSeek-R1-Distill-Qwen-1.5B} proxy evaluates predictive
entropy along each document. As described in Section~3.2, long
documents are processed with windows of 4096 tokens and a stride
of 2048 tokens; entropy estimates in overlapping regions are
averaged. High- and Low-Entropy candidates correspond to the
highest and lowest 5\% of token positions within a document,
respectively. For the Stochastic arm, each token position is
independently preselected with probability 0.05 using
deterministic document-level seeds.

Deduplication is performed separately for each arm. Within each
non-overlapping 512-token bucket, we retain the candidate with
the highest proxy entropy. Consequently, final anchors in the
Stochastic arm are not a uniformly random sample of token positions.
We then uniformly downsample the three deduplicated sets without
replacement to 435,049 anchor prompts per arm. Each prompt consists
of up to 1024 tokens preceding its selected anchor; the anchor token
itself is not included in the prompt.

\subsection{Distillation Implementation}
\label{app:distillation_details}

The OPMD student is \texttt{Qwen3-1.7B-Base}. We evaluate
\texttt{Qwen3-4B-Base} and \texttt{Qwen3-14B-Base} as frozen teachers.
For each anchor prompt, the student generates one on-policy
continuation of at most 512 tokens. At each generated position,
the distillation signal uses the student's top-16 next-token
candidates. Student probabilities are normalized within this
candidate set, while teacher log-probabilities are taken from
the teacher's full-vocabulary distribution. The individual
candidate contributions are used as advantage signals in the
PPO-based implementation, as described in Section~3.2.

The reported training configuration uses one epoch, a learning
rate of $1\times 10^{-6}$, and a reference-policy KL coefficient
of $\beta=0.005$. The different optimizer step counts in the
checkpoint tables should not be interpreted as different numbers
of source documents. Comparisons across training runs are shown
against raw training token volume in the main-text figures.

\subsection{Evaluation Protocol}
\label{app:evaluation}

We use the \texttt{gen\_vllm.py} generation script and
\texttt{grade.py} grading script from the public Rethinking OPD
implementation.\footnote{\url{https://github.com/Thinking-Space/Rethinking-OPD}}
For each problem in MATH-500 or Minerva, we generate 16 responses
and score each response using the same grading procedure. Let
$c_{ij}\in\{0,1\}$ indicate whether response $j$ to problem $i$
is graded correct, and let $N$ be the number of problems. We report
\begin{align}
\mathrm{pass@16}
&= \frac{1}{N}\sum_{i=1}^{N}
\mathbf{1}\!\left[\sum_{j=1}^{16}c_{ij}>0\right],\\
\mathrm{avg@16}
&= \frac{1}{16N}\sum_{i=1}^{N}\sum_{j=1}^{16}c_{ij}.
\end{align}
We multiply both quantities by 100 when reporting them as
percentages. Pass@16 measures the fraction of problems for which
at least one of the 16 responses is correct; avg@16 measures the
average correctness of an individual response over the same samples.

\section{Additional Checkpoint Results}
\label{app:checkpoint_results}

\setcounter{dbltopnumber}{2}
\renewcommand{\dbltopfraction}{0.95}

The main text presents training trajectories against raw training
token volume. Here we report the recorded checkpoint scores
underlying those trajectories. For each benchmark, pass@16 is the
percentage of problems with at least one correct response among
16 samples, while avg@16 is the average correctness across the
16 responses. All scores are percentages rounded to one decimal
place. Checkpoints from the same run are successive evaluations,
not independent training repetitions. Since optimizer step counts
differ across runs, step numbers should not be used as a direct
measure of matched training exposure.

\subsection{Teacher-Scale Comparison}
\label{app:teacher_scale_checkpoints}

Table~\ref{tab:opmd_step_results} compares High-Entropy OPMD
runs using Qwen3-4B-Base and Qwen3-14B-Base as teachers. Both
runs start from the same Qwen3-1.7B-Base student.

\begin{table*}[t]
\centering
\footnotesize
\caption{High-Entropy OPMD checkpoint results with 4B and 14B
teachers. The initial student is Qwen3-1.7B-Base. Each metric
is computed from 16 responses per question and reported in \%.}
\label{tab:opmd_step_results}
\begin{tabular*}{\linewidth}{@{\extracolsep{\fill}}lcccc}
\toprule
Checkpoint
& \multicolumn{2}{c}{MATH-500}
& \multicolumn{2}{c}{Minerva} \\
\cmidrule(lr){2-3}\cmidrule(lr){4-5}
& pass@16 & avg@16 & pass@16 & avg@16 \\
\midrule
\multicolumn{5}{l}{\textbf{Teacher: Qwen3-4B-Base}} \\
Student (initial) & 76.6 & 18.5 & 30.5 & 4.7 \\
Steps 100  & 80.6 & 27.5 & 33.8 & 6.7 \\
Steps 200  & 80.8 & 29.0 & 34.2 & 6.3 \\
Steps 300  & 81.8 & 29.0 & 35.3 & 6.9 \\
Steps 400  & 81.6 & 29.9 & 34.6 & 6.9 \\
Steps 500  & 82.0 & 31.3 & 34.2 & 6.9 \\
Steps 600  & 81.4 & 30.5 & 33.5 & 6.7 \\
Steps 700  & 81.4 & 30.5 & 34.9 & 7.0 \\
Steps 800  & 82.6 & 30.5 & 36.4 & 7.0 \\
Steps 900  & 82.2 & 30.5 & 33.8 & 6.9 \\
Steps 1000 & 81.8 & 31.2 & 35.7 & 7.0 \\
Steps 1100 & 82.2 & 29.9 & 35.3 & 7.2 \\
Steps 1200 & 81.4 & 30.9 & 33.1 & 6.7 \\
Steps 1300 & 80.4 & 31.3 & 33.5 & 6.6 \\
Steps 1400 & 82.4 & 29.8 & 37.5 & 6.8 \\
Steps 1500 & 82.0 & 30.2 & 34.9 & 6.8 \\
Steps 1600 & 83.4 & 30.5 & 34.9 & 6.9 \\
Steps 1699 (final) & 82.8 & 30.5 & 34.9 & 7.0 \\
\midrule
\multicolumn{5}{l}{\textbf{Teacher: Qwen3-14B-Base}} \\
Student (initial) & 76.6 & 18.5 & 30.5 & 4.7 \\
Steps 100  & 81.0 & 29.4 & 33.5 & 7.1 \\
Steps 200  & 82.4 & 30.3 & 33.8 & 7.1 \\
Steps 300  & 81.2 & 30.4 & 37.5 & 7.7 \\
Steps 400  & 80.4 & 30.5 & 36.0 & 7.1 \\
Steps 500  & 80.2 & 31.2 & 33.5 & 7.6 \\
Steps 600  & 80.8 & 29.7 & 35.7 & 7.6 \\
Steps 700  & 82.6 & 31.1 & 36.8 & 7.6 \\
Steps 800  & 81.2 & 31.9 & 38.2 & 8.1 \\
Steps 900  & 81.6 & 30.9 & 36.0 & 7.5 \\
Steps 1000 & 81.2 & 31.2 & 34.6 & 7.5 \\
Steps 1100 & 81.8 & 29.8 & 34.9 & 7.3 \\
Steps 1132 (final) & 82.4 & 29.9 & 36.0 & 8.0 \\
\bottomrule
\end{tabular*}
\end{table*}

\subsection{NTP Checkpoints}
\label{app:ntp_checkpoints}

Table~\ref{tab:ntp_step_results} reports the available
Qwen3-1.7B NTP checkpoints and the initial and final results
of the Llama-3.2-3B NTP pilot experiment. The two NTP
experiments use the same sampled set of original source
documents.

\begin{table*}[t]
\centering
\small
\caption{Available NTP checkpoint results on MATH-500 and
Minerva. Each metric is computed from 16 responses per
question and reported in \%.}
\label{tab:ntp_step_results}
\begin{tabular*}{\linewidth}{@{\extracolsep{\fill}}lcccc}
\toprule
Model and checkpoint
& \multicolumn{2}{c}{MATH-500}
& \multicolumn{2}{c}{Minerva} \\
\cmidrule(lr){2-3}\cmidrule(lr){4-5}
& pass@16 & avg@16 & pass@16 & avg@16 \\
\midrule
Qwen3-1.7B-Base (initial) & 76.6 & 18.5 & 30.5 & 4.7 \\
Qwen3-1.7B NTP, step 400 & 30.2 & 2.5 & 9.6 & 0.7 \\
Qwen3-1.7B NTP, step 800 & 17.8 & 1.5 & 8.8 & 0.7 \\
Qwen3-1.7B NTP, step 1200 & 17.6 & 1.3 & 6.6 & 0.4 \\
Qwen3-1.7B NTP, step 1600 & 20.0 & 1.5 & 8.1 & 0.6 \\
Qwen3-1.7B NTP, step 1902 (final) & 18.2 & 1.5 & 5.5 & 0.4 \\
\midrule
Llama-3.2-3B-Base (initial) & 11.2 & 1.1 & 7.3 & 1.0 \\
Llama-3.2-3B NTP (final) & 37.0 & 6.2 & 17.3 & 2.4 \\
\bottomrule
\end{tabular*}
\end{table*}

\subsection{Anchor-Selection Comparison}
\label{app:anchor_checkpoints}

Table~\ref{tab:anchor_step_results} lists all three
anchor-selection trajectories, including the High-Entropy
results from Table~\ref{tab:opmd_step_results}. Figure~4
compares the arms by raw training data volume.

\begin{table*}[t]
\centering
\footnotesize
\caption{Checkpoint trajectories for three anchor-selection
arms with the Qwen3-14B-Base teacher. Results within each arm
are listed in training order. Each metric is computed from
16 responses per question and reported in \%. Step numbers
identify checkpoints within a run and are not matched
training-token budgets across arms.}
\label{tab:anchor_step_results}
\begin{tabular*}{\linewidth}{@{\extracolsep{\fill}}lcccc}
\toprule
Checkpoint
& \multicolumn{2}{c}{MATH-500}
& \multicolumn{2}{c}{Minerva} \\
\cmidrule(lr){2-3}\cmidrule(lr){4-5}
& pass@16 & avg@16 & pass@16 & avg@16 \\
\midrule
\multicolumn{5}{l}{\textbf{High-Entropy}} \\
Student (initial) & 76.6 & 18.5 & 30.5 & 4.7 \\
Steps 100  & 81.0 & 29.4 & 33.5 & 7.1 \\
Steps 200  & 82.4 & 30.3 & 33.8 & 7.1 \\
Steps 300  & 81.2 & 30.4 & 37.5 & 7.7 \\
Steps 400  & 80.4 & 30.5 & 36.0 & 7.1 \\
Steps 500  & 80.2 & 31.2 & 33.5 & 7.6 \\
Steps 600  & 80.8 & 29.7 & 35.7 & 7.6 \\
Steps 700  & 82.6 & 31.1 & 36.8 & 7.6 \\
Steps 800  & 81.2 & 31.9 & 38.2 & 8.1 \\
Steps 900  & 81.6 & 30.9 & 36.0 & 7.5 \\
Steps 1000 & 81.2 & 31.2 & 34.6 & 7.5 \\
Steps 1100 & 81.8 & 29.8 & 34.9 & 7.3 \\
Steps 1132 (final) & 82.4 & 29.9 & 36.0 & 8.0 \\
\midrule
\multicolumn{5}{l}{\textbf{Low-Entropy}} \\
Student (initial) & 76.6 & 18.5 & 30.5 & 4.7 \\
Steps 100  & 78.8 & 21.6 & 31.6 & 6.0 \\
Steps 200  & 80.2 & 23.6 & 34.6 & 6.4 \\
Steps 300  & 78.0 & 24.3 & 31.6 & 5.5 \\
Steps 400  & 80.4 & 24.0 & 34.6 & 6.4 \\
Steps 500  & 77.0 & 24.3 & 32.0 & 5.4 \\
Steps 600  & 78.0 & 24.1 & 31.3 & 5.9 \\
Steps 700  & 78.0 & 23.2 & 33.1 & 6.0 \\
Steps 800  & 79.2 & 24.8 & 32.0 & 5.5 \\
Steps 900  & 80.6 & 25.3 & 31.6 & 6.1 \\
Steps 1000 & 80.4 & 24.5 & 33.5 & 5.9 \\
Steps 1100 & 79.0 & 25.1 & 34.9 & 5.6 \\
Steps 1200 & 79.6 & 24.6 & 32.4 & 5.4 \\
Steps 1300 & 79.8 & 24.3 & 32.7 & 5.7 \\
Steps 1400 & 79.0 & 24.5 & 34.9 & 5.5 \\
Steps 1500 & 80.2 & 25.2 & 33.1 & 6.1 \\
Steps 1600 & 77.0 & 24.6 & 36.4 & 6.5 \\
Steps 1699 (final) & 78.4 & 24.8 & 33.8 & 6.6 \\
\midrule
\multicolumn{5}{l}{\textbf{Stochastic Candidate Preselection}} \\
Student (initial) & 76.6 & 18.5 & 30.5 & 4.7 \\
Steps 100  & 80.4 & 26.2 & 33.5 & 7.4 \\
Steps 200  & 80.8 & 28.3 & 34.6 & 6.1 \\
Steps 300  & 81.6 & 28.4 & 34.9 & 6.6 \\
Steps 400  & 81.2 & 29.7 & 40.8 & 7.7 \\
Steps 500  & 82.6 & 28.6 & 34.6 & 7.4 \\
Steps 600  & 82.4 & 29.1 & 35.7 & 7.4 \\
Steps 700  & 80.8 & 29.0 & 34.2 & 7.3 \\
Steps 800  & 83.2 & 28.8 & 33.1 & 7.4 \\
Steps 900  & 81.6 & 28.9 & 33.5 & 6.7 \\
Steps 1000 & 80.6 & 29.4 & 34.6 & 6.9 \\
Steps 1100 & 82.6 & 29.3 & 33.1 & 6.6 \\
Steps 1132 (final) & 82.4 & 28.4 & 35.7 & 7.2 \\
\bottomrule
\end{tabular*}
\end{table*}